\documentclass{article}

\usepackage{arxiv}

\usepackage[utf8]{inputenc} 
\usepackage[T1]{fontenc}    
\usepackage{hyperref}       
\usepackage{url}            
\usepackage{booktabs}       
\usepackage{amsfonts}       
\usepackage{nicefrac}       
\usepackage{microtype}      
\usepackage{lipsum}		
\usepackage{graphicx}
\usepackage{natbib}
\usepackage{doi}

\usepackage{amssymb}
\usepackage{color}
\usepackage{multirow}
\usepackage[cmex10]{amsmath}
\usepackage{array}
\usepackage{algorithm}
\usepackage{algorithmic}

\title{Semi-parametric Makeup Transfer via \\ Semantic-aware Correspondence}

\author{ {\includegraphics[scale=0.06]{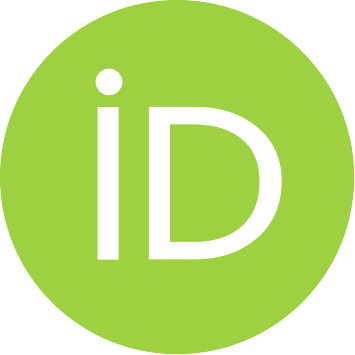}\hspace{1mm}Mingrui Zhu} \\
	State Key Laboratory of Integrated Services Networks \\
	Xidian University \\
	Xi’an, China \\
	\texttt{mrzhu@xidian.edu.cn} \\
	\And
	{\includegraphics[scale=0.06]{orcid.pdf}\hspace{1mm}Yun Yi} \\
	State Key Laboratory of Integrated Services Networks \\
	Xidian University \\
	Xi’an, China \\
	\texttt{yuny220@163.com} \\
	\And
	{\includegraphics[scale=0.06]{orcid.pdf}\hspace{1mm}Nannan Wang} \\
	State Key Laboratory of Integrated Services Networks \\
	Xidian University \\
	Xi’an, China \\
	\texttt{nnwang@xidian.edu.cn} \\
	\And
	{\includegraphics[scale=0.06]{orcid.pdf}\hspace{1mm}Xiaoyu Wang} \\
	The Chinese University of Hong Kong (Shenzhen) \\
	Shenzhen, China \\
	\texttt{fanghuaxue@gmail.com} \\
	\And
	{\includegraphics[scale=0.06]{orcid.pdf}\hspace{1mm}Xinbo Gao} \\
	Chongqing Key Laboratory of Image Cognition \\
	Chongqing University of Posts and Telecommunications \\
	Chongqing, China \\
	\texttt{gaoxb@cqupt.edu.cn} \\
}

\renewcommand{\shorttitle}{\textit{arXiv} Template}

\hypersetup{
pdftitle={Semi-parametric Makeup Transfer via Semantic-aware Correspondence},
pdfsubject={Makeup Transfer},
pdfauthor={Mingrui Zhu, Yunyi, Nannan Wang, Xiaoyu Wang, Xinbo Gao},
pdfkeywords={Makeup transfer, Semi-parametric, Semantic-aware correspondence, Generative adversarial networks},
}

\begin{document}
\maketitle

\begin{abstract}
	The large discrepancy between the source non-makeup image and the reference makeup image is one of the key challenges in makeup transfer. Conventional approaches for makeup transfer either learn disentangled representation or perform pixel-wise correspondence in a parametric way between two images. We argue that non-parametric techniques have a high potential for addressing the pose, expression, and occlusion discrepancies.To this end, this paper proposes a \textbf{S}emi-\textbf{p}arametric \textbf{M}akeup \textbf{T}ransfer (SpMT) method, which combines the reciprocal strengths of non-parametric and parametric mechanisms.The non-parametric component is a novel \textbf{S}emantic-\textbf{a}ware \textbf{C}orrespondence (SaC) module that explicitly reconstructs content representation with makeup representation under the strong constraint of component semantics. The reconstructed representation is desired to preserve the spatial and identity information of the source image while ``wearing'' the makeup of the reference image. The output image is synthesized via a parametric decoder that draws on the reconstructed representation. Extensive experiments demonstrate the superiority of our method in terms of visual quality, robustness, and flexibility. Code and pre-trained model are available at \url{https://github.com/AnonymScholar/SpMT}.
\end{abstract}


\section{Introduction}
\label{sec:Introduction}
Makeup transfer, aiming to apply cosmetics to a non-makeup face image by simulating a well-makeup one, has attracted tremendous interest and become an active topic in computer vision. Its applications can automatically improve users' facial appearance with chosen makeup examples and therefore save a great deal of manual work. 

\begin{figure*}[!htb] 
\centering
\includegraphics[width=1\linewidth]{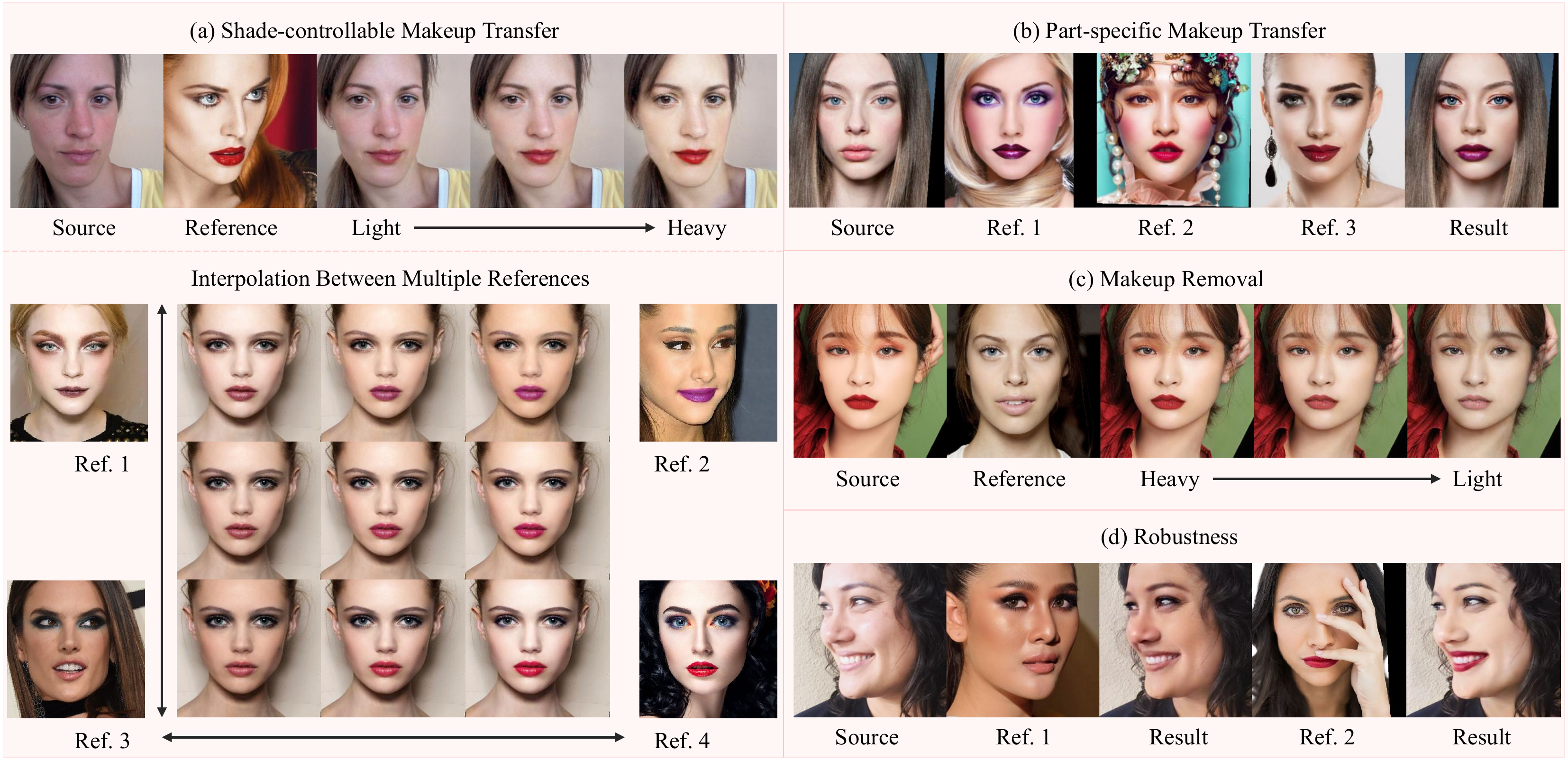}
\caption{The proposed SpMT method combines the reciprocal strengths of non-parametric and parametric techniques and has natural advantages in flexibility and robustness. (a) Shade-controllable makeup transfer by interpolation between non-parametric reconstructed representations. (b) Part-specific makeup transfer by selecting different parts from different reconstructed representations. (c) Makeup removal that does not require a bi-directional mapping, all in one parametric decoder. (d) Robustness to environmental variants.}
\label{fig:cover}
\end{figure*}

While early research endeavors~\citep{tong2007example}~\citep{guo2009digital}~\citep{li2015simulating} mainly leverage image processing approaches to accomplish this task, significant progress, however, is driven by recent deep learning-based methods~\citep{liu2016makeup}, the route of which has been much more prevalent since the booming of generative adversarial networks (GAN)~\citep{goodfellow2014generative}. Inspired by CycleGAN~\citep{zhu2017unpaired}, a domain-level image-to-image translation model, PairedCycleGAN~\citep{chang2018pairedcyclegan} and BeautyGAN~\citep{li2018beautygan} tackle makeup transfer as an instance-level exemplar-based translation problem, which opens a new route for following studies. Recent advances in disentangled representation~\citep{lee2018diverse} also energize many works~\citep{chen2019beautyglow, zhang2019disentangled, gu2019ladn, sun2020local, li2020disentangled} to learn the decomposition and recombination of makeup style representation and content representation. Currently, cutting-edge methods try to seek robust makeup transfer methods that can well against the environmental variants and adapt to the real-world scenario. Promising results have been reported in~\citep{huang2020real, jiang2020psgan, deng2021spatially, wan2021facial, lyu2021sogan}.

The aforementioned deep learning-based methods, despite demonstrating success, are all parametric models that represent the makeup procedure (\emph{i.e.} mapping function) via their weights. They forsake the competency to draw on the material provided by examples in exchange for highly efficient end-to-end capability. One obvious defect is that they always fail to cover up freckles or blemishes on the faces. In addition, this may not consist of the practices of the makeup artist. Imagine the real makeup procedure: the makeup artist picks up a powder and applies it evenly on the face, or applies lipstick with a certain color number to the lips. In this procedure, facial makeup comes from real makeup tools. Imitating this process, makeup transfer may also take the material of the corresponding component directly from the reference makeup image and apply it to the source image in a non-parametric manner.

In this paper, we combine the reciprocal advantages of parametric and non-parametric techniques, and present a \textbf{S}emi-\textbf{p}arametric \textbf{M}akeup \textbf{T}ransfer (SpMT) approach. The non-parametric makeup procedure is achieved through a newly proposed \textbf{S}emantic-\textbf{a}ware \textbf{C}orrespondence (SaC) module. This module elegantly establishes a semantic-aware correspondence between the corresponding components of the source non-makeup image and the reference makeup image in their feature pyramids. Whether in the training or testing process, the module obtains raw materials from the reference image through the semantic-aware correspondence and reconstructs the representation of the source image. The advantages of the non-parametric SaC module lie in its robustness against environmental variants like pose, expression, or occlusion discrepancies and its flexibility on controllable makeup transfer (as shown in Figure \ref{fig:cover}). The reconstructed representation is then processed by an image reconstruction network that produces an after-makeup image as output in a parametric manner. 

The contributions of this work are threefold:

1) A semi-parametric approach, which to our best knowledge, is the first to emphasize the potential of the non-parametric mechanism in the makeup transfer task.

2) A non-parametric SaC module accompanied with a cosmetic perceptual loss that establishes a semantic-aware correspondence between the non-makeup representation and makeup representation, which enables robustness and flexibility of the presented method.

3) Extensive experiments that demonstrate the superiority of the proposed method over state-of-the-arts and verify its capability to realize controllable and robust makeup transfer while yielding makeup images with satisfying quality.

\section{Related Work}
\label{sec:Related Work}

\subsection{Makeup Transfer}
Makeup transfer refers to such a kind of task that when given a pair of input images, transfer the specific makeup style from one of them (called reference image) to another (called source image) without destroying the face identity of the source input. It has been studied~\citep{tong2007example} \citep{guo2009digital} \citep{xu2013automatic} \citep{li2015simulating} for a decade before the ascendance of deep learning approaches. Rapid progress in deep convolutional neural networks, especially in generative adversarial networks, has inspired recent progress of makeup transfer. PairedCycleGAN~\citep{chang2018pairedcyclegan} introduced an asymmetric makeup transfer framework based on CycleGAN~\citep{zhu2017unpaired}. BeautyGAN~\citep{li2018beautygan} utilized a dual input/output generative adversarial network and a pixel-level histogram loss on local regions to fulfill instance-level makeup transfer. Based on Glow architecture~\citep{kingma2018glow}, BeautyGlow~\citep{chen2019beautyglow} decomposed the latent vectors into makeup vectors and facial identity vectors, and then invert the recombined vectors to the image domain. \citet{gu2019ladn} proposed LADN by incorporating local style discriminators, disentangling representation, and asymmetric loss functions into a cross-domain image translation network. Some studies \citep{zhang2019disentangled} \citep{sun2020local} \citep{li2020disentangled} also adopted the idea of disentangled representation learning. A recent line of studies~\citep{huang2020real} \citep{jiang2020psgan} \citep{deng2021spatially} \citep{wan2021facial}  \citep{lyu2021sogan} explored robust makeup transfer models that can well against the environmental variants. PSGAN~\citep{jiang2020psgan} introduced an attentive makeup morphing module based on an attention mechanism to realize partial, shade-controllable, and pose/expression robust makeup transfer. SCGAN~\citep{deng2021spatially} broke down the makeup transfer problem into a two-step extraction-assignment process and proposed a part-specific style encoder to map the makeup style into a component-wise style-code, which can eliminate the spatial misalignment problem. In this paper, we show that the proposed non-parametric SaC module, combined with the parametric decoder, also achieves excellent robustness and flexibility.

\subsection{Semi-parametric Studies}
The idea that combing the complementary strengths of parametric and non-parametric techniques has been cases in other research fields. \citet{qi2018semi} proposed a semi-parametric model to synthesize a photographic image from semantic layouts. The non-parametric component of their model is a memory bank of training image segments which used as raw materials for drawing on the semantic layout. In the field of image style transfer, \citet{liao2017visual} proposed a deep image analogy module that can find semantically-meaningful dense correspondences between two input images. \citet{gu2018arbitrary} proposed a feature reshuffle module that integrates global and local style losses which owns the advantages of both neural parametric and non-parametric methods. \citet{sheng2018avatar} and \citet{zhang2019multimodal} also adopted this idea and demonstrated its effectiveness. \citet{zhang2020cross} proposed a CoCosNet to jointly learn the cross-domain correspondence and the image translation. Inspired by these studies, we address the robust makeup transfer task by proposing a semi-parametric model. The non-parametric component is inspired by~\citep{chen2016fast}, but we make a complete improvement by the proposed SaC module for the specific characteristic of makeup transfer (For a discussion on the differences between image style transfer and makeup transfer, please refer to~\citep{li2018beautygan}).

\begin{figure*}[!tb]
\centering
\includegraphics[width=1\linewidth]{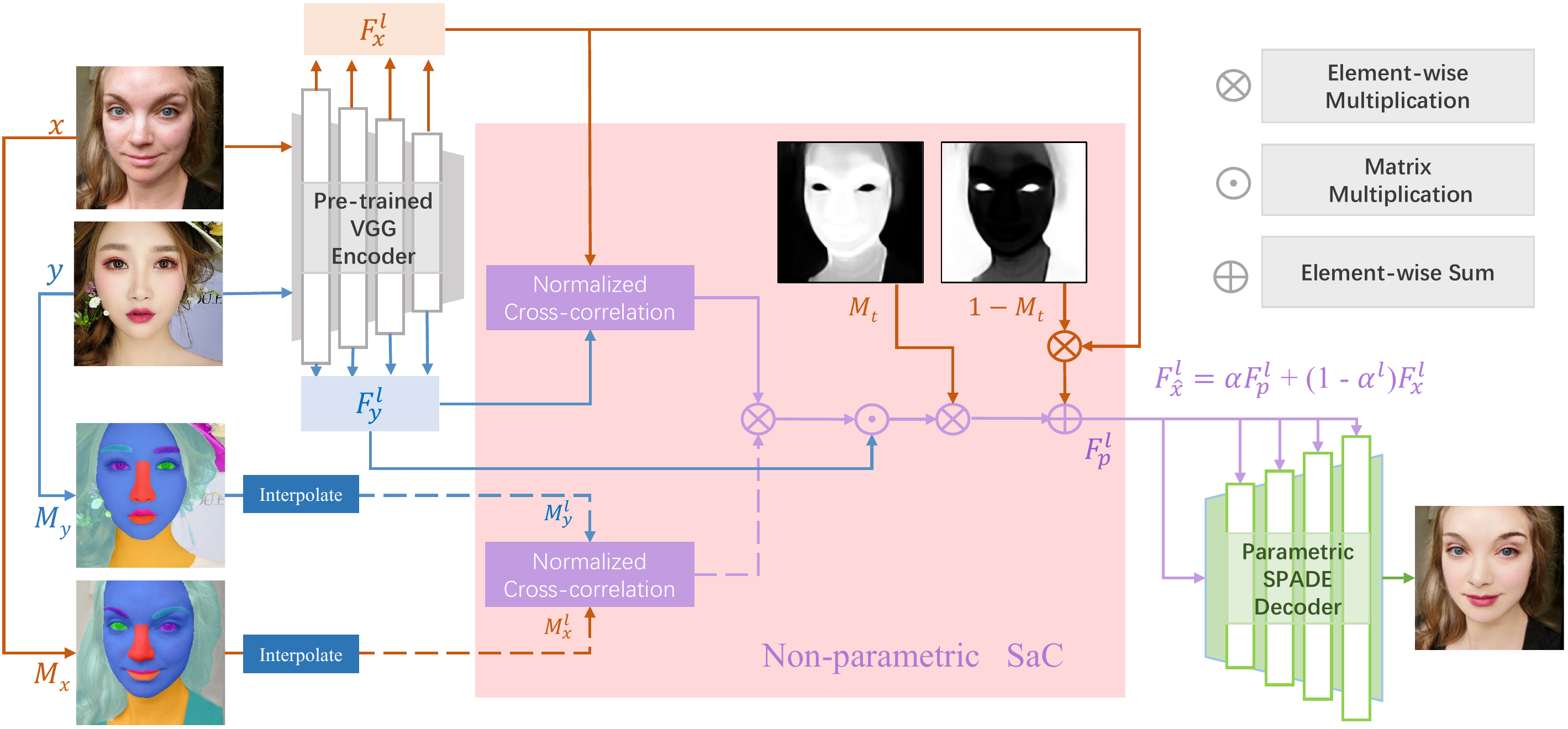}
\caption{The illustration of the SpMT framework. Pre-trained VGG-19 is used to encode the input images into multi-scale ($l$ denotes the scale number, $l=1,2,3,4$) deep representations. Then, the proposed SaC module reconstructs the pseudo representation in a non-parametric manner. The output image is synthesized via a parametric image reconstruction network that draws on the reconstructed representation. SpMT combines the reciprocal strengths of non-parametric and parametric mechanisms}
\label{fig:framework}
\end{figure*}

\section{Overview}
\label{sec:Overview}

Let $X = \{x_m \mid x_m \in X\}_{m = 1, \ldots M}$ denote the examples sampled from the non-makeup image distribution $\mathcal{P}_X$ and $Y = \{y_n \mid y_n \in Y\}_{n = 1, \ldots N}$ denote the examples sampled from the makeup image distribution $\mathcal{P}_Y$. For the task of makeup transfer, we aim to generate an after-makeup image $\hat{x}$ given a source non-makeup image $x$ and a reference makeup image $y$. That is, learning the mapping function: $\hat{x} = \mathcal{G}(x, y)$. The generated image $\hat{x}$ is desired to conform to the identity as $x$ while resembling the makeup style of the reference image $y$. 

Deep learning-based models typically represent the mapping function $\hat{x} = \mathcal{G}(x, y)$ via neural architectures with learnable weights, which belong to parametric models. Considering the advantage of non-parametric techniques for addressing the pose, expression, and occlusion discrepancies, we integrate the non-parametric module and parametric module in an end-to-end semi-parametric framework. 

The pipeline of SpMT is shown in Figure \ref{fig:framework}. We utilize the pre-trained VGG-19~\citep{simonyan2014very} to encode the source image $x$ and the reference image $y$ into their deep representations $F_{x}^{l}$ and $F_{y}^{l}$, where $l$ is the number of VGG-19 layers. Note that the parameters of the VGG-19 are frozen and will not change during training. Then, the proposed SaC module reconstructs pseudo-representation $F_{\hat{x}}^l$ given $F_{x}^{l}$ and $F_{y}^{l}$ in a non-parametric way. The reconstructed representation is desired to preserve the spatial distribution of $F_{x}^{l}$ and the cosmetic style of $F_{y}^{l}$. Unlike existing approaches that learn the disentanglement between content and style, the SaC directly takes raw materials from $F_{y}^{l}$ to decorate $F_{x}^{l}$, just like the makeup procedure performed by makeup artists. Note that the proposed SaC relies on facial mask $M_{x}^{l}$ of the source image and facial mask $M_{y}^{l}$ of the reference image that provides component semantics. SaC introduces semantic information into correspondence in a simple yet elegant way. The reconstructed representation $F_{\hat{x}}^l$ is used as input to a parametric image reconstruction network. This network synthesizes the final output image $\hat{x}$.

\begin{figure}[!tb]
\centering
\includegraphics[width=0.8\linewidth]{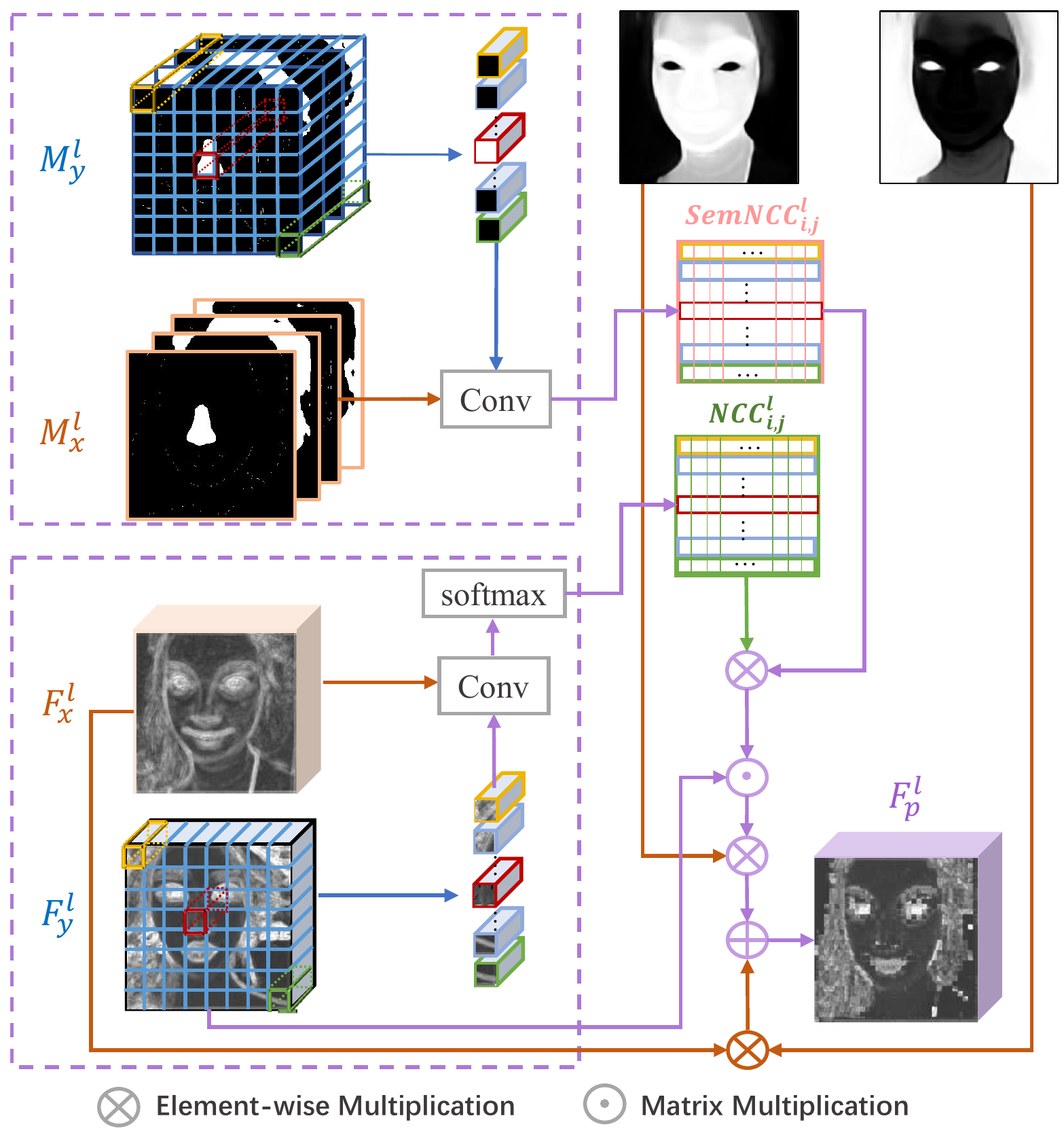}
\caption{The illustration of the SaC module. SaC computes two types of normalized cross-correlation: the normalized cross-correlation computed on deep representations, and the normalized cross-correlation computed on facial semantics (as shown in the dashed box above). By performing element-wise multiplication between them, we can obtain the semantic-aware correspondence, which is capable of reconstructing $F_{p}^l$ with makeup style patches only in the same semantic component}
\label{fig:sac}
\end{figure}

\section{Semantic-aware Correspondence}
\label{sec:Semantic-aware Correspondence}

Recent advances in attention mechanism~\citep{wang2018non} bring an efficient way to compute the response at a position as a weighted sum of the features at all positions, which establishes a dense pixel-wise correspondence between two matrices. However, the computational complexity on the spatial scale limits that it can only be calculated on the deep representations of small spatial size (generally the deep representations at the bottleneck). Actually, \citep{chen2016fast} provides another mechanism with the same effect but a different mean. Formally, let $\Psi(\cdot)$ denote the list of all local patches extracted from the deep representation. Each neural patch is indexed as $\Psi_{i}(\cdot)$ of the size $c \times k \times k$, where $c$ is the number of channels and $k$ is the spatial size of the patch. For each neural patch $\Psi_{i}(F_{x}^{l})$, best matching patch $\Psi_{NN(i)}(F_{y}^{l})$ can be found using normalized cross-correlation($NCC$) over all $N_y$ example patches in $\Psi(F_{y}^{l})$:
\begin{equation}
NCC_{i,j}^l(F_x^l, F_y^l) = 
\frac{\Psi_i(F_x^l) \odot \Psi_j(F_y^l)}
{\vert\Psi_i(F_x^l)\vert \odot \vert\Psi_j(F_y^l)\vert},
\end{equation}
\begin{equation}
NN^l(i):= \mathop{\arg\max}\limits_{j \in \{1,...,N_y\}} 
NCC_{i,j}^l(F_x^l, F_y^l).
\end{equation}

This process is actually done by executing convolution operation on $\Psi(F_{c}^{l})$ with custom convolution kernels $\Psi_{i}(F_{y}^{l})$ and computing the max response, all in a feed-forward convolutional layer. When the spatial size of the patch is set to 1 and computing the softmax instead of the max response, this module has the same function as the pixel-wise correspondence in the attention mechanism:
\begin{equation}
\Psi_i(F_{p}^l) = \sum_j \mathop{\rm{softmax}}\limits_{j} 
NCC_{i,j}^l(F_x^l, F_y^l)
\odot \Psi_j(F_y^l).
\end{equation}

This module breaks the restriction that the correspondence can only be established at the pixel level, so it can be applied to deep representation with a larger spatial size by using a large patch size.

However, this module is far from enough for makeup transfer. Since the correspondence suggests the matched reference patches strictly similar to the source patches, the reconstructed representation $\Psi_i(F_{p}^l)$ will be highly biased towards the source representation $F_{x}^{l}$ and only a limited portion of cosmetic patterns are transferred to $\Psi_i(F_{p}^l)$, especially when $F_{x}^{l}$ and $F_{y}^{l}$ are significantly discrepant from each other. The affect on the reconstructed image is that the makeup style of the reference image has been washed out (as shown in Figure \ref{fig:ablation_sac}). In particular, in the makeup transfer task, makeup style (\textit{e.g.} foundation, lipsticks, eye shadows) is more delicate and elaborate, which is closely related to local components. The global correspondence will destroy such local cosmetic patterns. 

Therefore, we present the SaC module that establishes correspondence at the semantic component level. The illustration of the SaC module is shown in Figure \ref{fig:sac}. In addition to the normalized cross-correlation computed on deep representations of the source image and the reference image (as shown in the dashed box below), we define a normalized cross-correlation on their facial semantics (as shown in the dashed box above):
\begin{equation}
SemNCC_{i,j}^l(M_x^l, M_y^l) = 
\frac{\Psi_i(M_x^l) \odot \Psi_j(M_y^l)}
{\vert\Psi_i(M_x^l)\vert \odot \vert\Psi_j(M_y^l)\vert}.
\end{equation}

Note that $M_{x}^{l}$ and $M_{y}^{l}$ both have multiple channels. Each channel corresponds to one component (\textit{e.g.} skin, lips, eyes). Each value in all channels is specified as 0 or 1, indicating that this position does not belong to or belongs to a specific component respectively. For example, a vector $[0,1,0,...,0]$ of a spatial position indicates that this position belongs to the facial skin. Therefore, the response (\textit{i.e.} dot product) of the vectors in two positions with the same semantic will be 1 and the response of the vectors in two positions with different semantics will be 0. By performing element-wise multiplication between $NCC_{i,j}^l$ and $SemNCC_{i,j}^l$, we can obtain the semantic-aware correspondence, which is capable of reconstructing $\Psi_i(F_{p}^l)$ with makeup style patches only in the same semantic component:
\begin{equation}
\begin{aligned}
\Psi_i(F_p^l) = 
&\sum_j \mathop{\rm{softmax}}\limits_{j} 
NCC_{i,j}^l(F_x^l, F_y^l) \\
&\otimes SemNCC_{i,j}^l(M_x^l, M_y^l) \odot \Psi_j(F_y^l).
\end{aligned}
\end{equation}

By stitching patches of $\Psi_i(F_{p}^l)$ together, $F_{p}^l$ is obtained. In the makeup transfer task, we only need to transfer the makeup of some specific components, so we can replace the parts in the reconstructed representation $F_{p}^{l}$ that do not need to be transferred with the source representation through mask $M_{t}$:
\begin{equation}
F_{p}^{l} = M_{t} \otimes F_{p}^{l} + (1 - M_{t}) \otimes F_{x}^{l}.
\end{equation}

$M_{t}$ is a binary mask, where all positions with a value of 1 constitute components that need makeup transfer and positions with a value of 0 constitute components that do not need makeup transfer.

\begin{figure*}[!tb]
\centering
\includegraphics[width=1.0\linewidth]{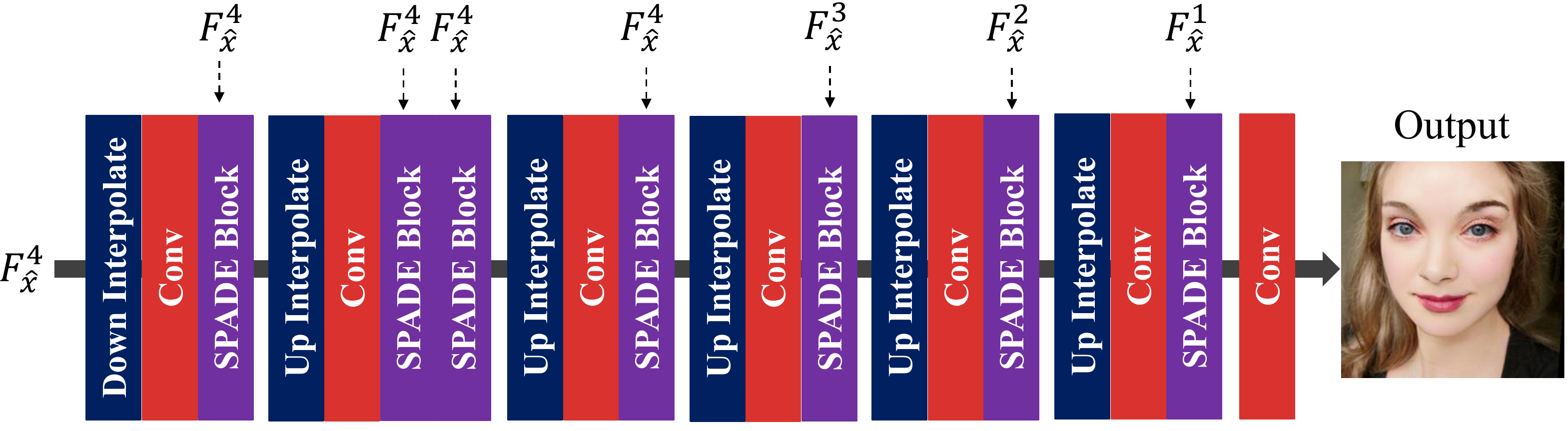}
\caption{The architecture of the SPADE decoder.}
\label{fig:spade_decoder}
\end{figure*}

In practice, we find that directly using $F_{p}^{l}$ will produce slight artifacts on a few generated images. Therefore, before inputting the reconstructed representation $F_{p}^{l}$ into the parametric image reconstruction network, we add different proportions of content features to $F_{p}^{l}$ on different scales to get a more stable performance:
\begin{equation}
F_{\hat{x}}^l = SaC(x,y) = \alpha^l F_{p}^{l} + (1 - \alpha^l) F_{x}^{l},
\label{con:feat_reconstruct}
\end{equation}
where $F_{\hat{x}}^l$ denotes the final reconstructed representation obtained via the SaC module $SaC(x,y)$.

Since the proposed SaC module works in a non-parametric manner and relies on facial masks to establish the semantic-aware correspondence, it can well against environmental variants like large pose, expression, and occlusion discrepancies. In addition, SaC can also achieve high flexibility like shade-controllable and part-specific by controlling the combination of $F_{\hat{x}}^l$ and $F_x^l$ through facial masks and weights.

\section{Image Reconstruction Network}
\label{sec:Image Reconstruction Network}

\subsection{SPADE Decoder} 
We require a feed-forward parametric model $\mathcal{G}$ which takes as input the multi-scale reconstructed representation $F_{\hat{x}}^l$ to reconstruct the after-makeup output $\hat{x}$. To this end, we design an image reconstruction network that consists of several spatially-adaptive de-normalization (SPADE)~\citep{park2019semantic} blocks. Each scale of the reconstructed representation $F_{\hat{x}}^l$ is used as input of each SPADE block to learning the modulation parameters $\alpha^l$ and $\beta^l$. The produced $\alpha^l$ and $\beta^l$ are multiplied and added to the normalized activation of the output of the previous layer element-wise. Formally, given the activation $F^l \in \mathfrak{R}^{c_l \times h_l \times w_l}$ before the $l^{th}$ block, we inject the $F_{\hat{x}}^l$ through:
\begin{equation}
\alpha_{h_l,w_l}^l(F_{\hat{x}}^l) \times 
\frac{F_{c_l,h_l,w_l}^l-\mu_{h_l,w_l}^l}{\sigma_{h_l,w_l}^l}. 
+ \beta_{h_l,w_l}^l(F_{\hat{x}}^l).
\end{equation}

The overall parametric reconstruction can be formulated as:
\begin{equation}
\hat{x} = \mathcal{G}(F_{\hat{x}}^l;\theta_\mathcal{G}), 
\end{equation}
where $\theta_G$ denotes the learnable parameter.

The architecture of the SPADE decoder is shown in Figure \ref{fig:spade_decoder}. The reconstructed representation $F_{\hat{x}}^4$ at $Relu\_4\_1$ is firstly down-interpolated and then gradually up-interpolated to reconstruct the output image. The image reconstruction network comprises 7 SPADE blocks. Each SPADE block is introduced to de-normalize the activations with learned parameters $\gamma$ and $\beta$ from a reconstructed representation $F_{\hat{x}}^l$.  We use $bilinear$ interpolation and $kernel - 3 \times 3 - stride - 1 - padding - 1$ convolution in this architecture. 

\subsection{Loss Function}

\subsubsection{Adversarial Loss}
We use the Least Squares GAN~\citep{mao2017least} to provide adversarial loss for the parametric SPADE decoder, which can help generate plausible natural-looking images with high perceptual quality. For the mapping function $\mathcal{G}$ and its discriminator $\mathcal{D}$, we express the adversarial loss as:
\begin{equation}
\begin{aligned}
\mathcal{L}_{adv} &= \mathbb{E}_{x, y}[(\mathcal{D}_Y(x, y))^2] \\
&+ \mathbb{E}_{x, y}[(1 - \mathcal{D}_{Y}(x, \mathcal{G}(SaC(x,y))))^2].
\end{aligned}
\end{equation}

\subsubsection{Content Loss}
Since some parts (\textit{e.g.} teeth, hair, etc.) in the makeup transfer task do not need to be transferred, a content loss is introduced to keep the non-transferred parts unchanged: 
\begin{equation}
\mathcal{L}_{con} = \mathbb{E}_{x, y}[\Vert (\mathcal{G}(SaC(x,y) - x) \otimes (1 - M_t) \Vert_2].
\end{equation}

\subsubsection{Cosmetic perceptual Loss}
The reconstructed representation ${F_{\hat{x}}^l}$ by the proposed non-parametric SaC module can be further used as pseudo-ground-truth to guide the training of $\mathcal{G}$. Therefore, we propose a cosmetic perceptual loss to substitute the raw perceptual loss \citep{johnson2016perceptual} that is commonly used in existing methods. Let $\varPhi^l(\cdot)$ denote the pyramid representation derived from the pre-trained VGG-19, the cosmetic perceptual loss is computed by: 
\begin{equation}
\mathcal{L}_{cos} = \mathbb{E}_{x, y}[\sum_l \Vert \varPhi^l(\mathcal{G}(SaC(x,y))) - F_{\hat{x}}^l \Vert_2].
\end{equation}

Since the reconstructed representation can preserve the spatial distribution of the source representation and the cosmetic style of the reference representation, this loss can force the generated image to preserve the spatial distribution of the source image and the cosmetic style of the reference image.

\subsubsection{Style Loss}
A certain proportion of global style loss can help improve the overall style of the output image, we use mean and variance of the representation in VGG-19 layers to model global style:
\begin{equation}
\begin{aligned}
\mathcal{L}_{sty} &= \mathbb{E}_{x, y}[ \sum_l\Vert\mu(\varPhi^l(\mathcal{G}(SaC(x,y))))-\mu(\varPhi^l(y))\Vert_2 \\
& + \sum_l\Vert\sigma(\varPhi^l(\mathcal{G}(SaC(x,y))))-\sigma(\varPhi^l(y))\Vert_2].
\end{aligned}
\end{equation}

\subsubsection{Makeup Loss}
We utilize the makeup loss proposed by~\citep{li2018beautygan} to further guide the makeup style of lips, eye shadows, and face regions:
\begin{equation}
\begin{aligned}
\mathcal{L}_{makeup} &= \mathbb{E}_{x, y}[ \mathcal{G}(SaC(x,y)) - HM(x,y) \Vert_1 ],
\end{aligned}
\end{equation}
where $HM(\cdot,\cdot)$ denotes the histogram matching in lips, eye shadows, and face regions and the output of $HM(x,y)$ has the makeup style of $y$ while preserving the identity of $x$.

\subsubsection{Total Loss}
By combining the above losses, we can achieve our full loss:
\begin{equation}
\begin{aligned}
\mathcal{L}_{total} &= \lambda_{adv} \mathcal{L}_{adv} + \lambda_{makeup} \mathcal{L}_{makeup} \\
&+ \lambda_{cos} \mathcal{L}_{cos} + \lambda_{sty} \mathcal{L}_{sty} + \lambda_{con} \mathcal{L}_{con},
\end{aligned}
\end{equation}
where $\lambda_{adv}$, $\lambda_{makeup}$, $\lambda_{cos}$, $\lambda_{sty}$, and $\lambda_{con}$ are the weights to balance the multiple objectives.

\section{Experiments}
\label{sec:Experiments}

\begin{figure}[!tb]
\centering
\includegraphics[width=0.8\linewidth]{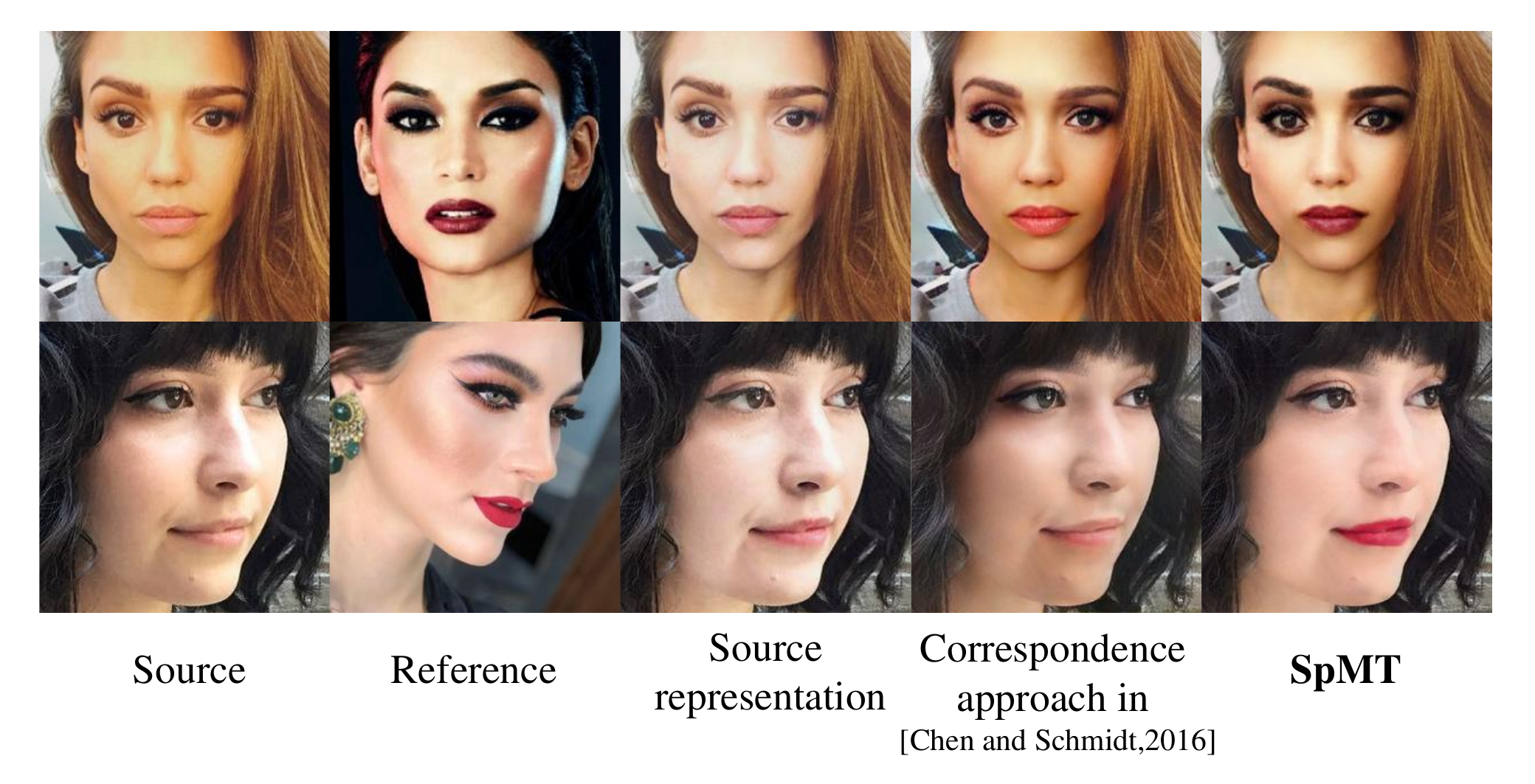}
\caption{Ablation study of the proposed SaC module.}
\label{fig:ablation_sac}
\end{figure}

\begin{figure}[!tb]
\centering
\includegraphics[width=0.8\linewidth]{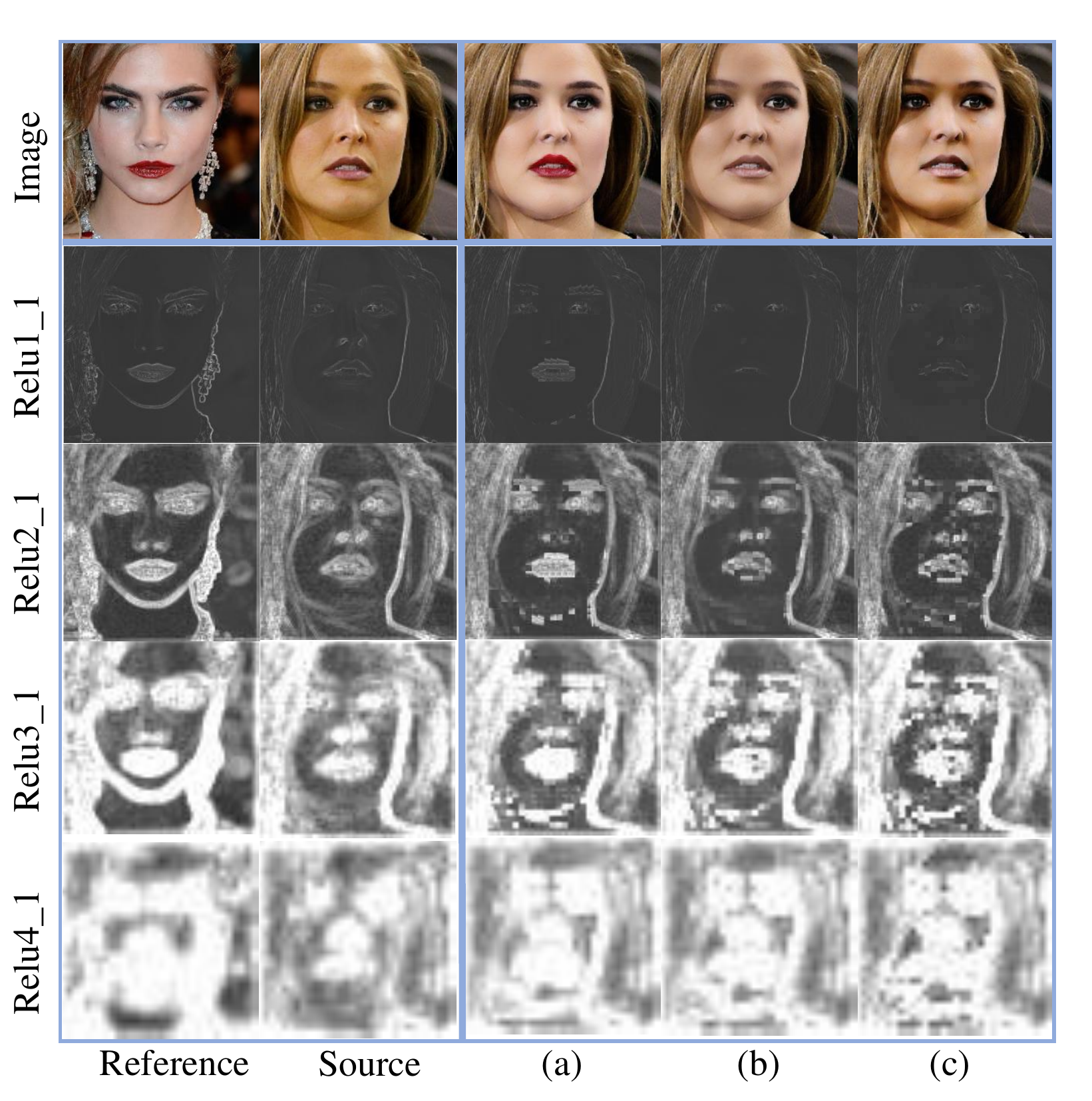}
\caption{Reconstructed multi-scale features and transferred images
with different correspondence methods. (a) The proposed semantic-aware correspondence. (b) Weighted correspondence. (c) Nearest patch correspondence \citep{chen2016fast}. To facilitate the comparison of generated makeup, we show the input reference image in the upper left corner.}
\label{fig:ablation_cos_and_scale}
\end{figure}

\subsection{Experimental Settings}
In this section, we first discuss the experimental settings. Then, we conduct an ablation study to quantify the contribution of different configurations to the overall effectiveness. Further, we evaluate the flexibility and robustness of the proposed model. Finally, we qualitatively and quantitatively compare our results with state-of-the-art methods.

\subsubsection{Implementation details.}
Feature activations extracted from $Relu\_1\_1$, $Relu\_2\_1$, $Relu\_3\_1$ and $Relu\_4\_1$ layers of the pre-trained VGG-19~\citep{simonyan2014very} are used as deep representations of input images. To establish semantic-aware correspondence,  we use $8 \times 8$ patch size for $Relu\_1\_1$, $4 \times 4$ patch size for $Relu\_2\_1$, $2 \times 2$ patch size for $Relu\_3\_1$ and $1 \times 1$ patch size for $Relu\_4\_1$. The convolution stride for each layer is equal to the patch size so that there is no overlap between the extracted patches. 

We set $\lambda_{adv}=1$, $\lambda_{makeup}=1$, $\lambda_{cos}=5$, $\lambda_{sty}=10$, $\lambda_{con}=100$. RMSProp optimizer with default parameters is utilized for optimization. The learning rate is fixed at 0.0002 and the batch size is set to 1. We scale the size of the input images to $256 \times 256$ and normalize the pixel value to the interval $[-1, 1]$ before putting them into the model. To acquire facial masks, we use the face parsing model provided in~\citep{faceparsing}, which is trained on the CelebAMask-HQ database~\citep{lee2020maskgan}.

\subsubsection{Dataset}
Our model was trained on the Makeup Transfer (MT) dataset~\citep{li2018beautygan} which contains 3834 images. It consists of 2719 makeup images and 1115 non-makeup images including variations in race, poses, expression, and background clutter. We use the same setting as \citet{li2018beautygan} that randomly select 100 non-makeup images and 250 makeup images for testing. The rest of the images are used for training.

\begin{figure*}[!tb]
\centering
\includegraphics[width=1.0\linewidth]{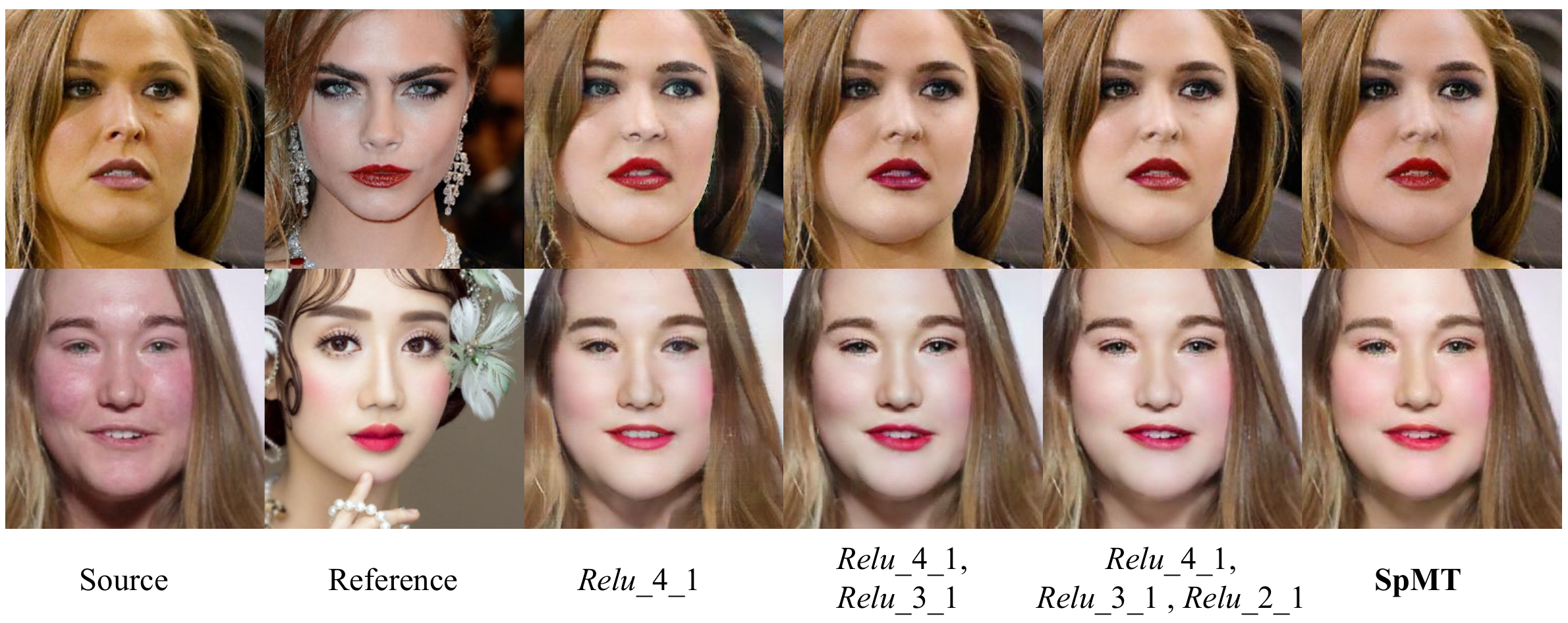}
\caption{Ablation study about the effect of different scales of the reconstructed representation on the final performance.}
\label{fig:ablation_scale}
\end{figure*}

\begin{figure}[!tb]
\centering
\includegraphics[width=0.8\linewidth]{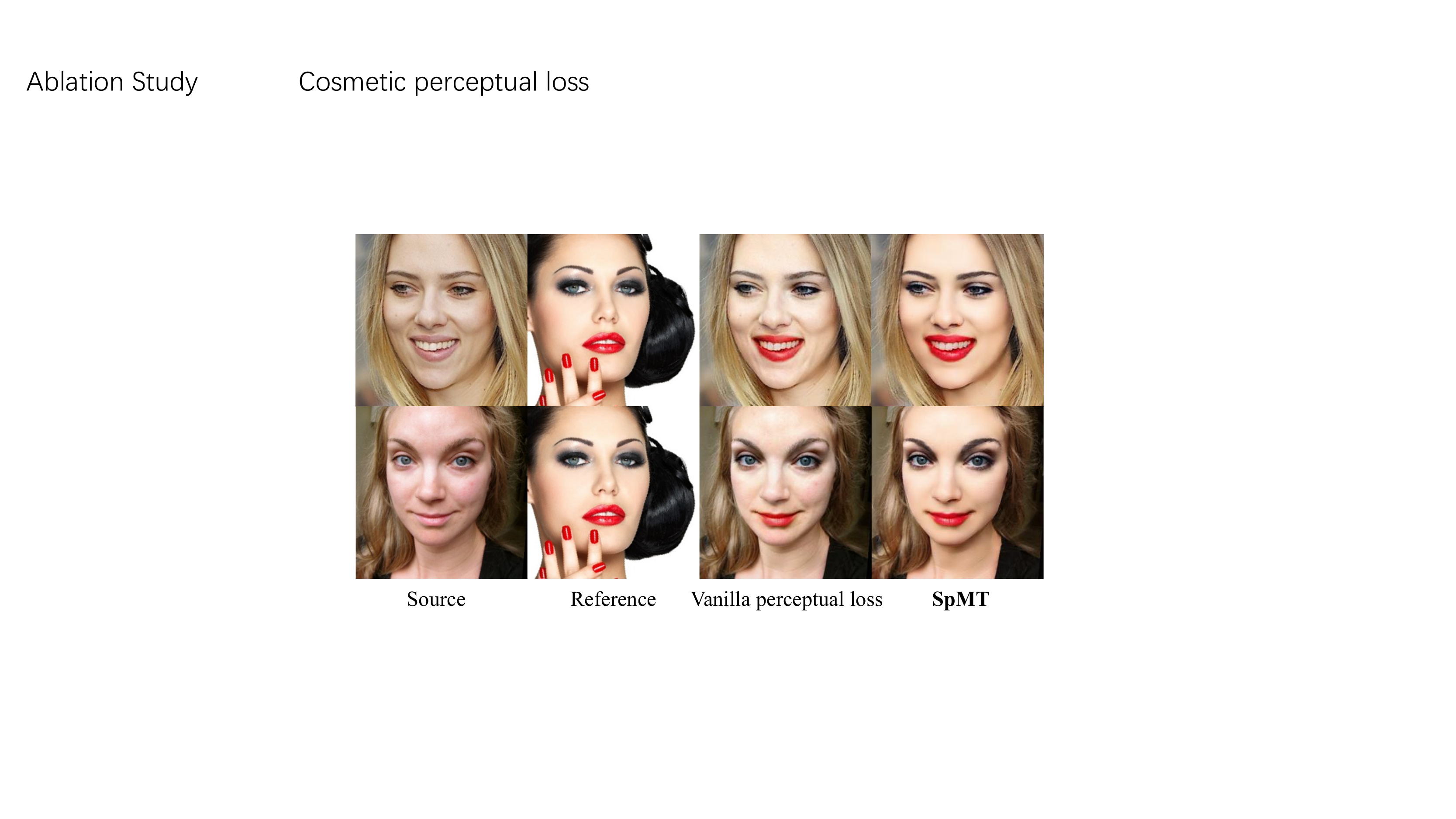}
\caption{Ablation study of cosmetic perceptual loss.}
\label{fig:ablation_cos}
\end{figure}

\begin{figure}[!tb]
\centering
\includegraphics[width=0.75\linewidth]{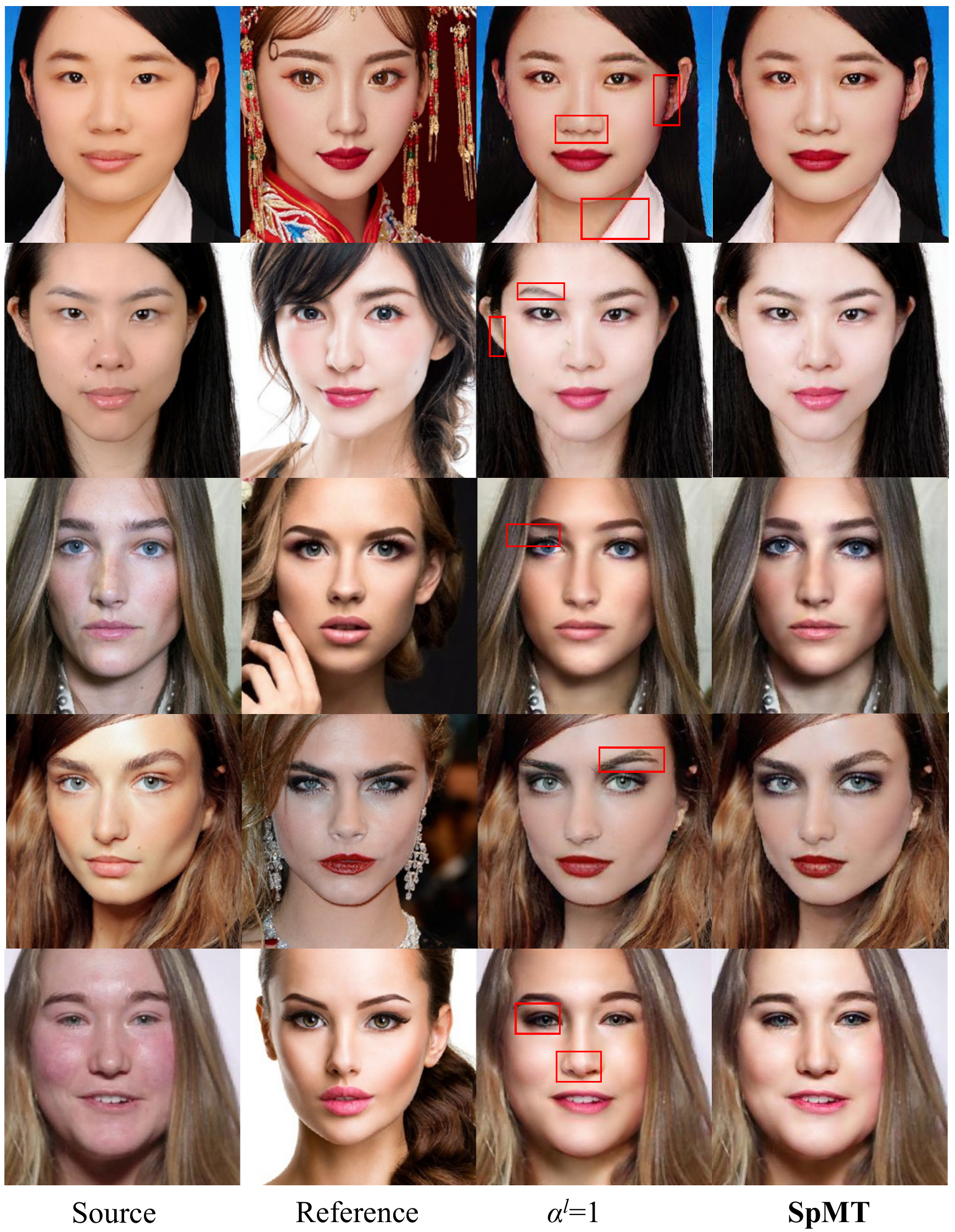}
\caption{Ablation study of $\alpha^l$.}
\label{fig:ablation_alpha}
\end{figure}

\subsection{Ablation Study}

Four contributions conduce to the overall efficacy.

\subsubsection{Non-parametric semantic-aware correspondence}
To verify the superiority of the SaC module, we replace it with the other two modules to observe the change of performance. Figure \ref{fig:ablation_sac} shows the results. When directly using the source representation instead of the reconstructed representation as to the input of the SPADE decoder, the generated image has almost no makeup style of the reference image. But its overall style is changing towards the reference image with the help of the parametric decoder. When using the correspondence approach provided by \citep{chen2016fast} instead of the SaC module, the makeup effect of the generated image will also be reduced. The global correspondence will wash away the local cosmetic patterns when the source image and the reference image are significantly discrepant from each other. The proposed SaC module establishes correspondence at the semantic component level to explicitly retain cosmetic patterns, thus producing the best results. 

Figure \ref{fig:ablation_cos_and_scale} visualizes the reconstructed multi-scale features and transferred images with different correspondence modules. As we can see, compared with the other two correspondence modules, features reconstructed by the proposed SaC module can better preserve the component content and transfer more reasonable makeup style. 

\subsubsection{Multi-scale}
The effect of different scales of the reconstructed representation on the final performance is shown in Figure \ref{fig:ablation_scale}. Using only the reconstructed representation of $Relu\_4\_1$ will cause some structural distortion on the generated image. Add $Relu\_3\_1$ and $Relu\_2\_1$ can alleviate this problem and present synthetic image with reasonable structure. However, when zoomed in, the problem that they lack reasonable style patterns in the hair, eye shadow, and foundation area will be exposed. SpMT that uses all layers of reconstructed representations can achieve the best results.

Figure \ref{fig:ablation_cos_and_scale} further presents the visualization results of the reconstructed representation of different scales. Reconstructed features of $Relu\_4\_1$ and $Relu\_3\_1$ contain the main structure and identity information of the image. Reconstructed features of $Relu\_2\_1$ and $Relu\_1\_1$ possess more texture information of the image. Therefore, by introducing the multi-scale mechanism, the proposed method also has the nice property that synthesizing the image from coarse-to-fine.

\subsubsection{Cosmetic perceptual loss}

To verify the efficacy of the cosmetic perceptual loss, we replace this loss with vanilla perceptual loss and observe the performance change. Figure \ref{fig:ablation_cos} shows the results. As we can see, compared with the vanilla perceptual loss, the cosmetic perceptual loss can transfer a more delicate and more consistent makeup style from the reference image to the source image.

\subsubsection{Setting of $\alpha^l$}

As mentioned in Sec. \ref{sec:Semantic-aware Correspondence}, we do not directly use the reconstructed representation $F_{p}^{l}$ as input to the SPADE decoder because it will produce slight artifacts on a few generated images. Therefore, we add different proportions of content features to $F_{p}^{l}$ on different scales (as shown in Eq. \ref{con:feat_reconstruct}) to get the final reconstructed representation $F_{\hat{x}}^l$ with more stable performance.Note that the $\alpha^l$ for each layer has a different value. We empirically set $\alpha^4=0.1$, $\alpha^3=0.2$, $\alpha^2=0.4$, and $\alpha^1=1$ to gradually reduce the proportion of content representation. The input of the first four blocks are reconstructed representations ($\alpha^4=0.1$) at $Relu\_4\_1$. The input of the 5th block is the reconstructed representation ($\alpha^3=0.2$) at $Relu\_3\_1$. The input of the 6th block is the reconstructed representation ($\alpha^2=0.4$) at $Relu\_2\_1$. The input of the 7th block is the reconstructed representation ($\alpha^1=1$) at $Relu\_1\_1$. The ablation results are shown in Figure \ref{fig:ablation_alpha}. When $\alpha^l$ is set to 1, which means directly using $F_{p}^{l}$ as input to the SPADE decoder, some artifacts will be produced on the output images. By fusing content representation in a certain proportion, the SpMT shows more stable performance.

\begin{figure}[!tb]
\centering
\includegraphics[width=0.8\linewidth]{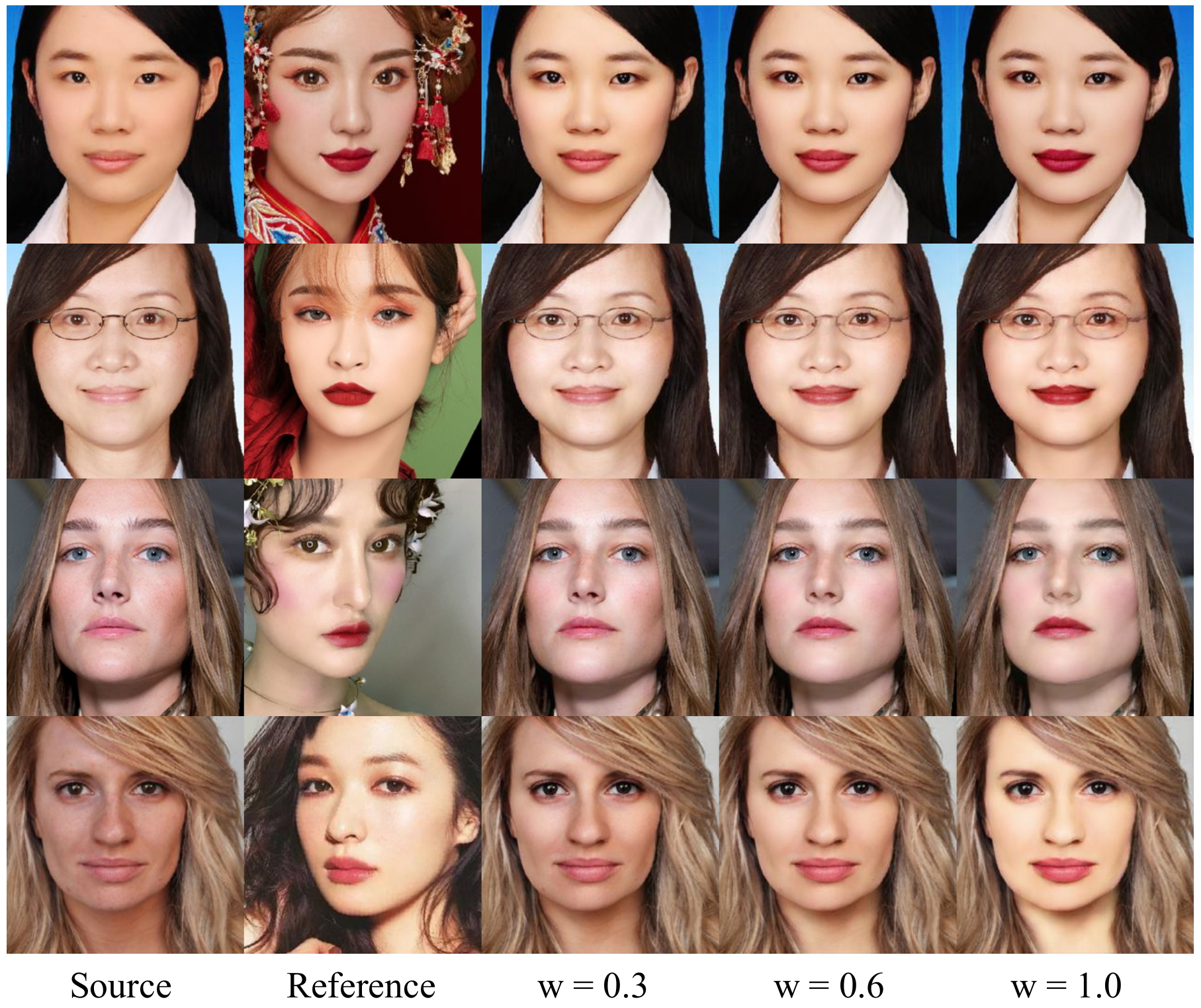}
\caption{Results of interpolated makeup styles. Adjusting the shade of makeup.}
\label{fig:controllable_shade1}
\end{figure}

\begin{figure}[!tb]
\centering
\includegraphics[width=0.7\linewidth]{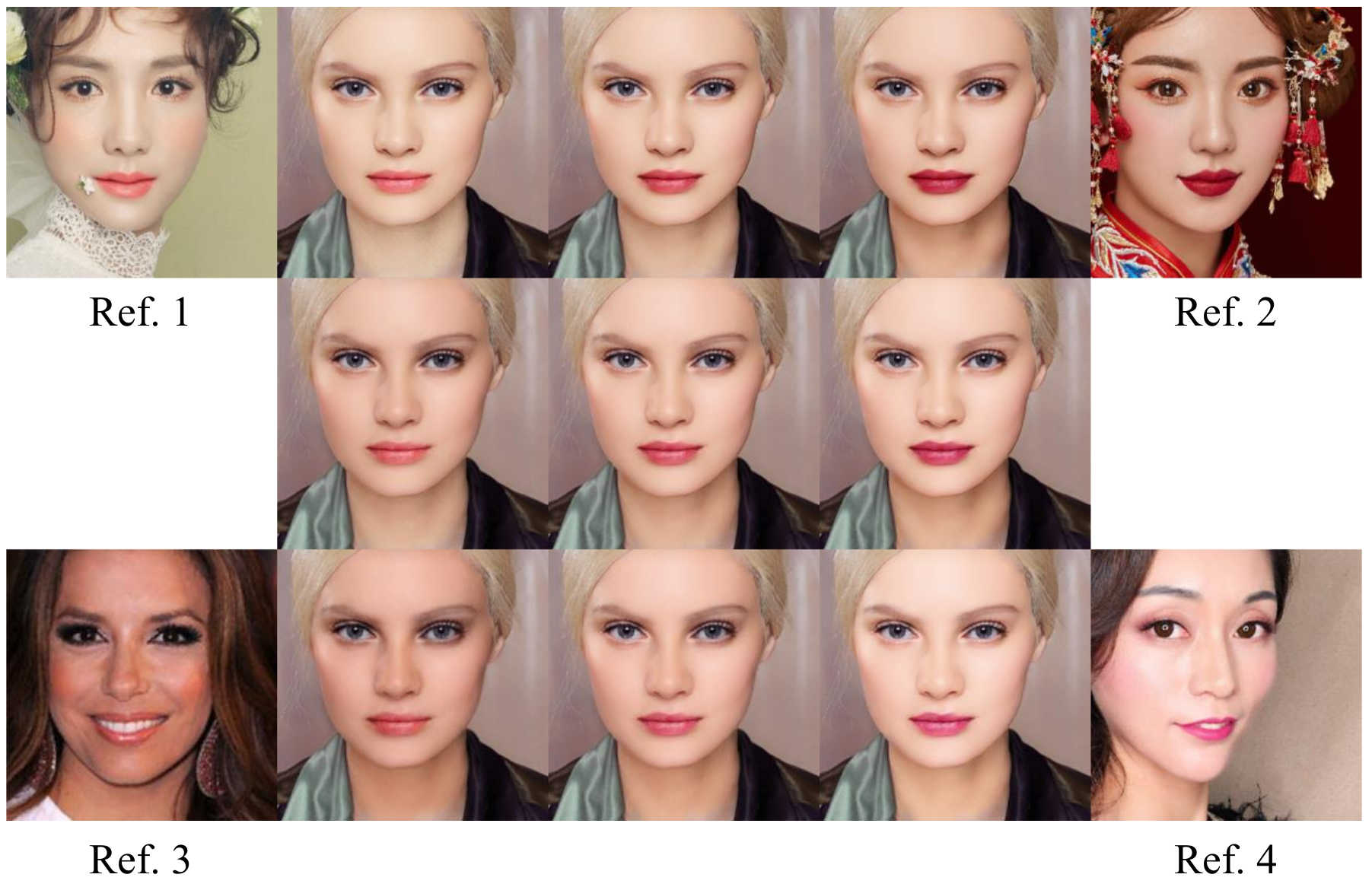}
\caption{Results of interpolated makeup styles. Interpolation between four references.}
\label{fig:controllable_shade2}
\end{figure}

\subsection{Flexibility and Robustness}

Since SpMT possesses a non-parametric module SaC, it has natural advantages in controllability and robustness.

\subsubsection{Shade-controllable transfer}
Shade-controllable transfer can be realized simply by interpolating the combined coefficient ($w \in [0, 1]$) of content representation and reconstructed representation:
\begin{equation}
R_s = w F_{\hat{x}}^l + (1-w) F_{x}^{l}
\end{equation}
By inputting the re-combined representation $R_s$ (with $w$ from 0 to 1) into the learned parametric SPADE decoder, we can gradually change the makeup style from the source image to the reference image.

SpMT also supports fusing the makeup styles from multiple reference images with a linear interpolation:
\begin{equation}
R_s = w_1 F_{\hat{x}-{ref1}}^l + w_2 F_{\hat{x}-ref2}^l + \ldots + w_k F_{\hat{x}-refk}^l
\end{equation}
where $w_1 + w_2 + \ldots + w_k = 1$. By changing the value of $w_k$, we can fuse multiple makeup styles from different reference images. 

Figure \ref{fig:cover} (a) shows the interpolated makeup transfer results and the makeup transfer results produced by mixing four reference images, which demonstrates the superb flexibility of SpMT. Figure \ref{fig:controllable_shade1} and Figure \ref{fig:controllable_shade2} present more examples of shade-controllable makeup transfer. Our method can adjust the shade of makeup and interpolation between multiple references.

\subsubsection{Part-specific transfer}
\begin{figure}[!tb]
\centering
\includegraphics[width=0.8\linewidth]{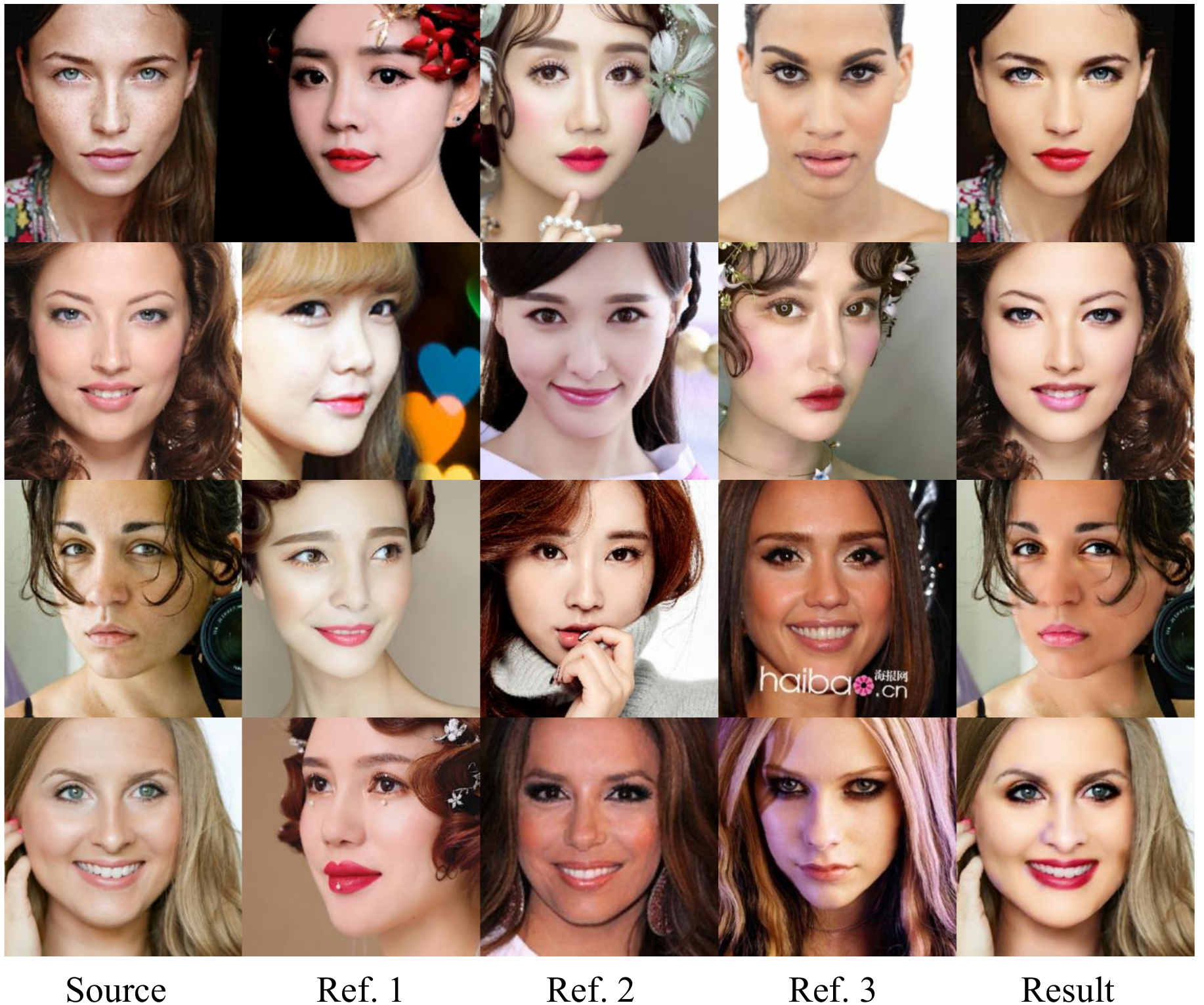}
\caption{Results of part-specific makeup transfer. Transfer makeup styles from different parts of different references.}
\label{fig:controllable_part}
\end{figure}

Since SpMT uses facial masks to distinguish semantic components and provide semantic constraints, it can easily perform part-specific makeup transfer by choosing any specific part from any reference image. For example, if we want to transfer the makeup styles of the lip, eye-shadow, and foundation from $Ref1$, $Ref2$, and $Ref3$ respectively, we can manipulate their reconstructed representation by:
\begin{equation}
\begin{aligned}
R_s &= M_l \otimes F_{\hat{x}-{Ref1}}^l + M_e \otimes F_{\hat{x}-Ref2}^l + M_f \otimes F_{\hat{x}-Ref3}^l \\
&+ (1-M_t) \otimes F_{x}^{l}
\end{aligned}
\end{equation}
where $M_t$ is a binary mask indicating where a transfer is required, $M_l$, $M_e$, and $M_f$ are binary masks indicating the lip, eye, and skin components. The result presented in Figure \ref{fig:cover} (b) successfully transferred the lip color, eye shadow, and foundation style from Ref.1, Ref.2, and Ref.3 respectively. Figure \ref{fig:controllable_part} shows some examples of part-specific makeup transfer. Our method can transfer makeup styles from different parts of different references to a single output image. 

\subsubsection{Makeup removal}
\begin{figure}[!tb]
\centering
\includegraphics[width=0.8\linewidth]{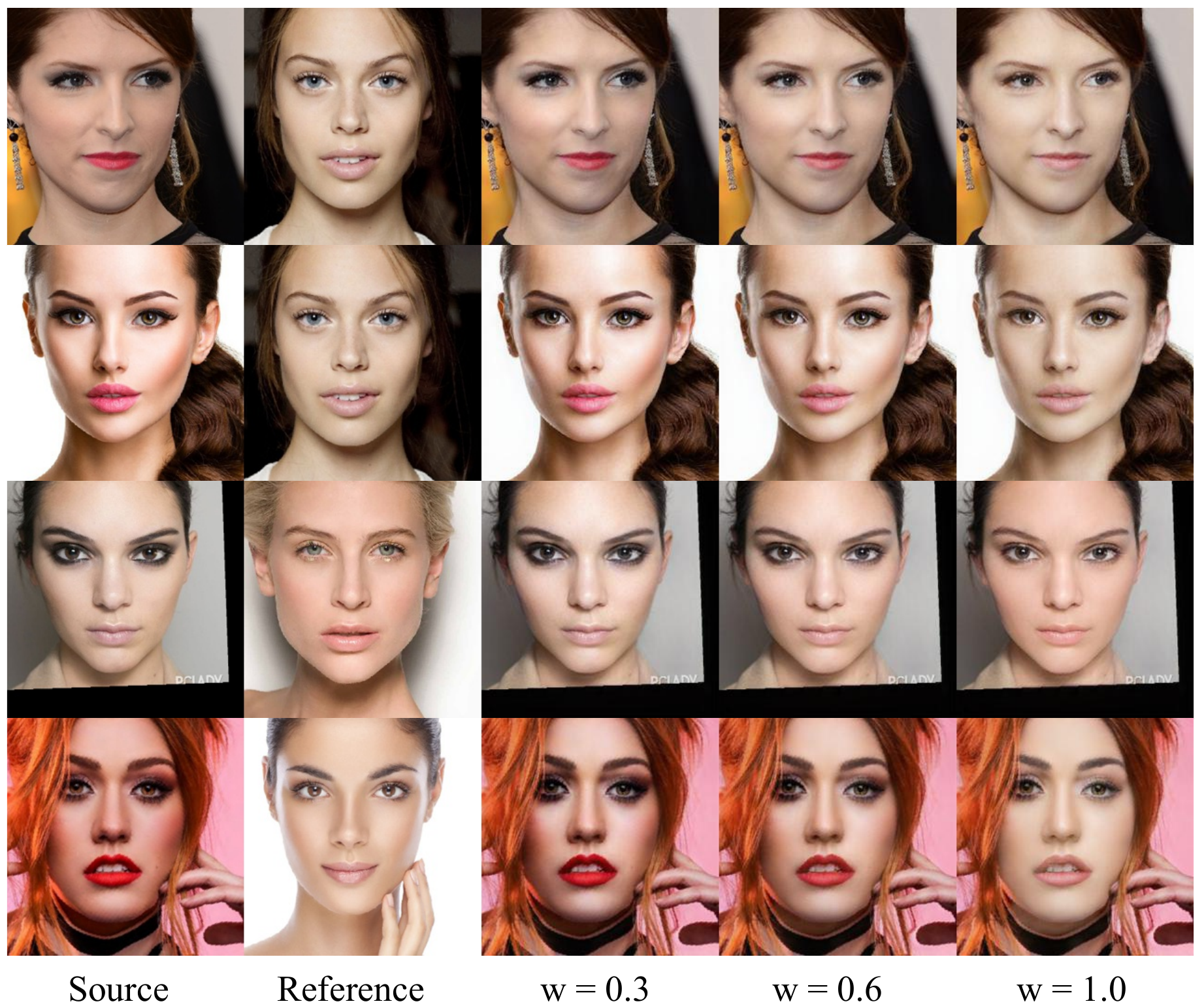}
\caption{Results of makeup removal.}
\label{fig:controllable_removal}
\end{figure}

Thanks to the powerful SaC module, the proposed SpMT method can achieve makeup removal by directly taking the non-makeup image as the reference image and the makeup one as the source image. SaC can reconstruct the non-makeup representation while keeping the identity of the makeup image. Note that SpMT does not require a bi-directional mapping as existing methods and performs image reconstruction using one SPADE decoder. Figure \ref{fig:cover} (c) and Figure \ref{fig:controllable_removal} show makeup removal results from heavy to light.

\subsubsection{Robustness}
\begin{figure*}[!tb]
\centering
\includegraphics[width=1.0\linewidth]{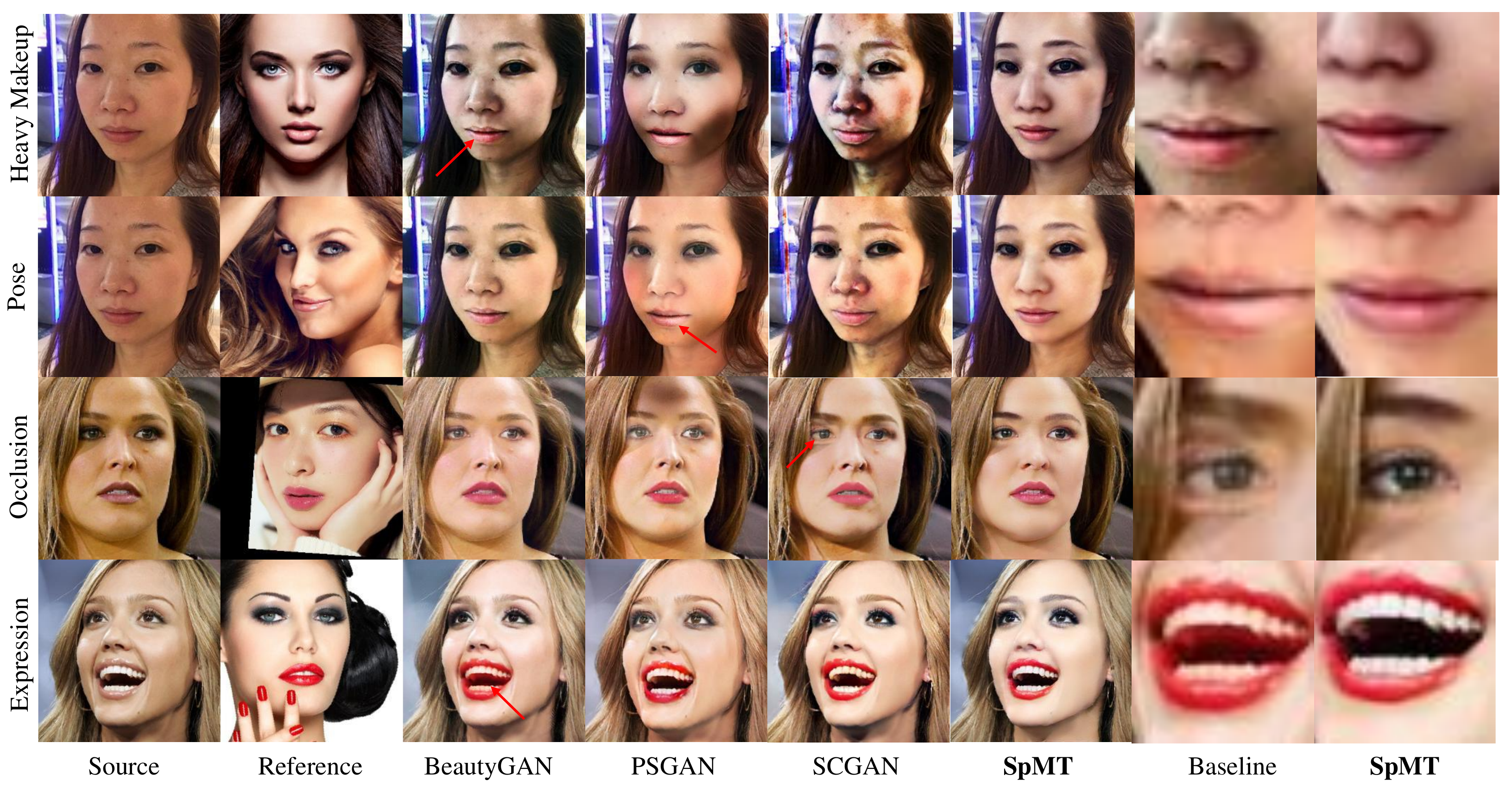}
\caption{Qualitative comparison with existing methods on faces with a large makeup style, pose, occlusion, and expression variants. The last two columns are the enlarged view of the area marked by the red arrow in baseline and the enlarged view of the corresponding part of SpMT.}
\label{fig:comparison_wild}
\end{figure*}

As mentioned before, SpMT has excellent robustness to the environmental variants owed to the advantage of the SaC module. Results presented in Figure \ref{fig:cover} (d) and Figure \ref{fig:comparison_wild} demonstrate this superiority.

\begin{figure*}[!tb]
\centering
\includegraphics[width=1.0\linewidth]{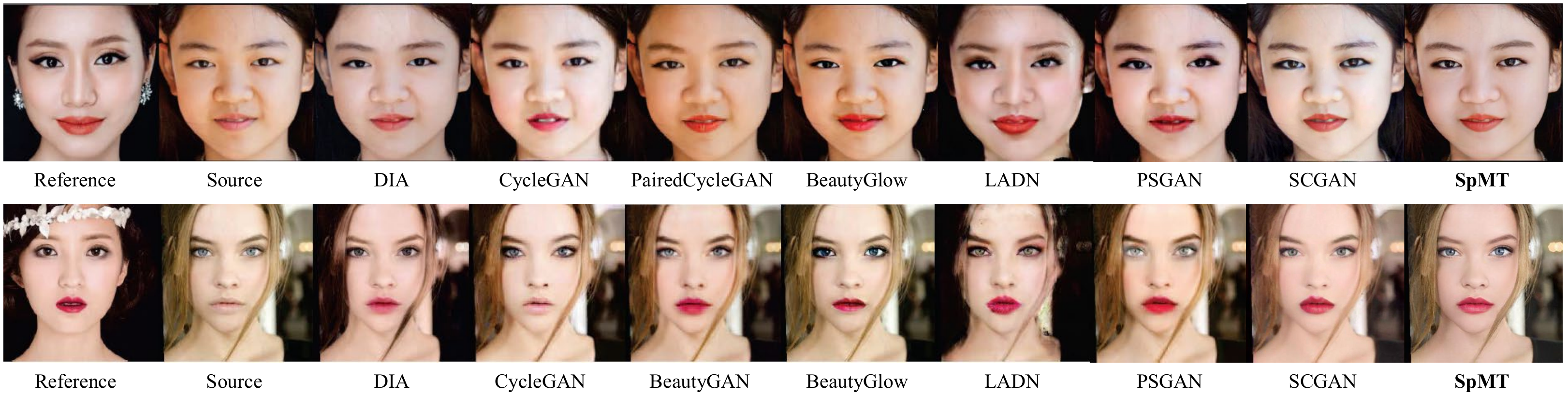}
\caption{Qualitative comparison with existing methods on normal faces without a large pose, expression, and occlusion variants.}
\label{fig:comparison_normal}
\end{figure*}

\subsection{Comparison with Prior Arts}

To validate the efficacy of the proposed method, we compare it with two general domain transfer methods CycleGAN~\citep{zhu2017unpaired} and DIA~\citep{liao2017visual} as well as six state-of-the-art makeup transfer methods PairedCycleGAN~\citep{chang2018pairedcyclegan}, BeautyGAN~\citep{li2018beautygan}, BeautyGlow~\citep{chen2019beautyglow}, LADN~\citep{gu2019ladn}, PSGAN~\citep{jiang2020psgan} and SCGAN~\citep{deng2021spatially}. Since the code of BeautyGlow and PairedCycleGAN are not available, we directly copy the results of BeautyGlow and PairedCycleGAN from their papers following the strategy used in BeautyGlow. The results of other methods are obtained by running their official code with the default configuration.

\subsubsection{Qualitative comparison}
Figure \ref{fig:comparison_normal} shows the qualitative comparison of SpMT with other state-of-the-art methods on normal faces without large pose, expression, and occlusion variants. Since DIA performs patch matching globally and has no semantic control, its results fail to maintain the content of components that do not need makeup transfer and fail to transfer a correct makeup style. Without the reference image, CycleGAN can not perform instance-level makeup transfer that imitates the specified makeup style. Results of PairedCycleGAN and BeautyGlow also have incorrect makeup styles that are inconsistent with the reference image, such as the color of the foundation. Severe artifacts affect the overall appearance of the results produced by LADN. BeautyGAN, PSGAN, and SCGAN perform well on these examples but still suffer from some minor defects. 

\begin{figure*}[!tb]
	\centering
	\includegraphics[width=1.0\linewidth]{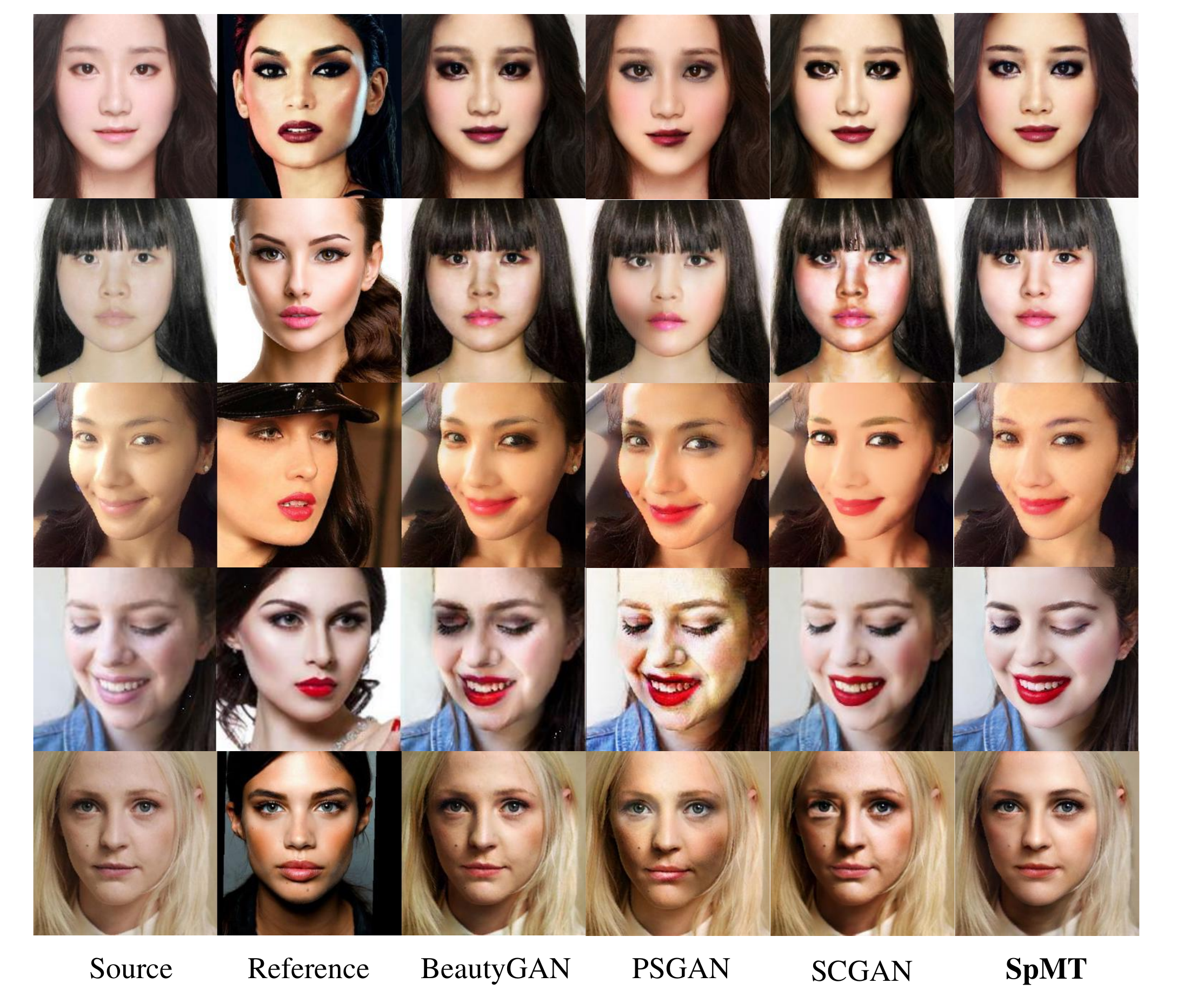}
	\caption{Additional examples of qualitative comparisons with existing methods}
	\label{fig:comparison_wild2}
\end{figure*}

To verify the robustness against the environmental variants such as pose, expression, and occlusion discrepancies, we further conduct a comparison with BeautyGAN, PSGAN, and SCGAN on some more challenging examples. The results are shown in Figure \ref{fig:comparison_wild}. As shown in the first row, when the makeup style of the input reference image is very thick, the results of the other three existing methods are affected, with uneven cosmetic, artifacts or significant noise. If there is occlusion on the face of the reference image, the result obtained by PSGAN appears obvious shadow at the corresponding position (the forehead of the face shown in the third row), which affects the appearance of the image. When the expression of the input source image is exaggerated as shown in the fourth row, cosmetic patterns of the lips appeared inside the mouth of the image generated by BeautyGAN, which damages the structure of the component. Another defect of these three methods is that the foundations of their results are not well imitating the reference images. The wrinkles and blemishes on the face are not well covered up (see the last two column of the first row for example), which is an unwanted result when people are beautifying themselves with cosmetics. Compared with cutting-edge methods, SpMT can produce better results with a more uniform and exquisite cosmetic style and fewer artifacts. For more results, please refer to Figure \ref{fig:comparison_wild2}.

\subsubsection{Quantitative comparison}
To compare the results of our method with the existing methods more objectively, we further conduct quantitative experiments.

For a particular makeup style, the images generated by a better makeup transfer algorithm should have a more similar distribution with its input reference images. Therefore, we calculated the Fr\'{e}chet Inception Distance (FID) \citep{fid} between the synthetic images and the reference images to measure the effect of makeup style transfer. Each row of Table \ref{tab:FID scores} shows the FID score for each method on one kind of makeup style. Every score was calculated using 300 images generated by randomly selected 10 non-makeup images and 30 with-makeup images. Results show that our model achieves the lowest FID score in the transfer of three makeup styles, which proves that our method can get results that more consistent with the makeup style of the reference image.

After makeup transfer, the content and structure information of the input image should not be changed. So we use Structural Similarity Index Metric (SSIM) \citep{wang2004image} to evaluate the degree of content maintenance. A higher score corresponds to better content preservation. From the results shown in Table \ref{tab:SSIM scores}, we can see that the SSIM score of our method is higher than other methods, which demonstrates the superiority of our method in maintaining the structure of the input source image.

\begin{table}[!tb]
\centering
\scalebox{1.0}{
\begin{tabular}{@{}ccccc@{}}
\toprule
Method  & BeautyGAN & PSGAN & SCGAN & SpMT           \\ \midrule 
Western & 155.1     & 156.7 & 154.5 & \textbf{141.3} \\
vFG     & 175.7     & 162.8 & 159.0 & \textbf{156.4} \\
vHX     & 140.6     & 131.8 & 136.7 & \textbf{119.2} \\ \bottomrule
\end{tabular}}
\caption{The experiment results of makeup style transferring}
\label{tab:FID scores}
\end{table}

\begin{table}[!tb]
\centering
\scalebox{1.0}{
\begin{tabular}{c|cccc} 
\toprule
Method & BeautyGAN & PSGAN & SCGAN & SpMT          \\ \midrule
SSIM$\uparrow$  & 0.87      & 0.80  & 0.81  & \textbf{0.89} \\ \bottomrule
\end{tabular}}
\caption{The experiment results of identity preserving}
\label{tab:SSIM scores}
\end{table}

\subsubsection{User study}
\begin{figure}[!tb]
\centering
\includegraphics[width=0.6\linewidth]{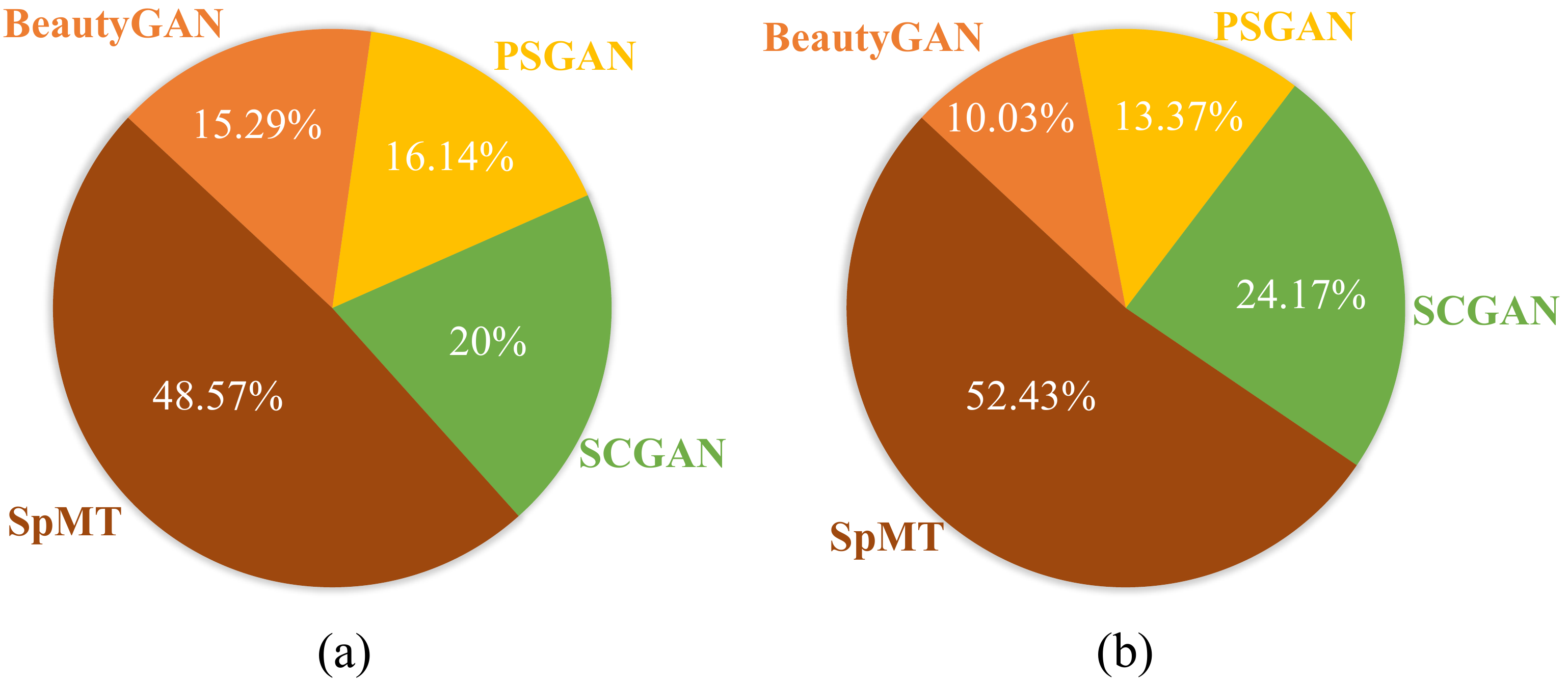}
\caption{User study results. (a) Ratio of votes on the results of well-aligned images. (b) Ratio of votes on the results of images with a large pose, expression, and occlusion variants.}
\label{fig:user_study}
\end{figure}

We conducted a user study to evaluate our algorithm against 3 state-of-the-art makeup transfer methods: BeautyGAN, PSGAN, and SCGAN. It consists of two parts. In the first part, we randomly selected 20 non-makeup images and 20 makeup images that are well-aligned. Link of the first part: \url{https://www.wenjuan.com/s/UZBZJvJrWR/#}. In the second part, we randomly selected 20 non-makeup images and 20 makeup images which are faces with a large pose, expression, and occlusion variants. Link of the second part: \url{https://www.wenjuan.com/s/aMRRJf2/#}. After performing makeup transfer between these images, we obtained 400 after-makeup images for each method in each part. During the test, 15 groups of results in part 1 and 15 groups of results in part 2 were randomly presented to the participant. For each group, the placement order of the 4 synthetic images was
shuffled and participants were asked to vote for their favorite result. Finally, we collected 1400 votes from 35 participants and calculated the percentage of votes for each method. The results are shown in Figure \ref{fig:user_study}. SpMT obtains the highest ratio of votes on both two parts.

\section{Conclusion}
\label{sec:Conclusion}
In this paper, we propose a semi-parametric approach for makeup transfer. Unlike existing methods that represent the mapping function from non-makeup images to makeup ones totally via parametric neural networks, the proposed method has a novel non-parametric semantic-aware correspondence module that directly gets raw materials from the reference image representation and decorates the source image representation. By combing the reciprocal strengths of parametric and non-parametric modules, the proposed approach achieves a great degree of flexibility and robustness. SpMT has a limitation that the non-parametric part is somewhat computational-expensive. It takes about 1 second to generate an image using a single Nvidia 3090 GPU. Properly reducing the number of layers of reconstruction representations can alleviate this problem. We will also explore more efficient techniques in the future.

\bibliographystyle{unsrtnat}
\bibliography{references}  






\end{document}